# A Framework of Customer Review Analysis Using the Aspect-Based Opinion Mining Approach

Subhasis Dasgupta[1], Jaydip Sen[2]
Dept. of Data Science
Praxis Business School
Kolkata, India
emails: [1]subhasis@praxis.ac.in, [2]jaydip.sen@acm.org

*Abstract*—Opinion mining is the branch of computation that deals with opinions, appraisals, attitudes, and emotions of people and their different aspects. This field has attracted substantial research interest in recent years. Aspect-level (called aspect-based opinion mining) is often desired in practical applications as it provides detailed opinions or sentiments about different aspects of entities and entities themselves, which are usually required for action. Aspect extraction and entity extraction are thus two core tasks of aspect-based opinion mining. his paper has presented a framework of aspect-based opinion mining based on the concept of transfer learning. on real-world customer reviews available on the Amazon website. The model has yielded quite satisfactory results in its task of aspect-based opinion mining.

## I. INTRODUCTION

Opinion mining is the branch of computation that deals with opinions, appraisals, attitudes, and emotions of people and their different aspects. This field has attracted substantial research interest in recent years. Aspect-level (called aspect-based opinion mining) is often desired in practical applications as it provides detailed opinions or sentiments about different aspects of entities and entities themselves, which are usually required for action. Aspect extraction and entity extraction are thus two core tasks of aspect-based opinion mining. This paper presents a framework of aspect-based opinion mining on real-world customer reviews available on the Amazon website.

According to Lui and Zhang, "opinion is a concept covering sentiment, evaluation, appraisal or attitude held by a person" [1]. Aspects and entities are topics included in a text document. Hu & Liu described the analysis of these topics and entities as feature-based sentiment analyses [2]. Aspect or entity-based analysis identifies the target of the opinion. It is a fine-grained approach to text analysis.

Consumer reviews hold important information about the product features and their level of acceptance by the end users. However, they fall under unstructured data, and capturing information from unstructured data is way too difficult than analyzing tabular data using established tools. It is quite common that when a reviewer gives a review, it contains some of the other aspects on which the review is focused at. Hence, it is quite important to understand the aspect(s) about which the review is more concentrated. The associated opinions are also important to be extracted to understand the degree of acceptance (or rejection) of the aspects attached to them. There are prior research works done to extract the aspect-opinion phrases such as discussed in the Related Works section. In most of the research, the researchers attempted to train the models with a significantly large number of annotated texts or by creating rules. However, when we use transfer learning, it is possible to train a custom model using a very small set of data. The current study focuses on this method because, through this process, over a period, a more powerful and accurate model can be created using the concept of semi-supervised learning. The current study shows that, first, a custom aspect-based opinion extraction model can be built using the concept of transfer learning. Second, the model can be fine-tuned rather quickly using a small sample of data. And third, the model can perform better than a random guess even when the reviews contain lots of grammatical errors.

The paper is structured as follows. In Section II, some existing works on portfolio design and stock price prediction are discussed briefly. Section III presents a description of the methodology followed in the work systematically. Section IV presents the performance results of the proposed approach. Section V concludes the paper while identifying some future directions of research.

## II. RELATED WORK

In recent years, aspect-based opinion mining and sentiment analysis have attracted substantial interest from the research community. In this section, some of the important works in these areas are mentioned.

Chinsha & Joseph proposed a method based on dependency parsing for syntactical analysis [3]. The proposed approach by the authors computes the aggregate score of opinion words based on the aspect table and SentiWordNet for opinion mining. The model was evaluated on restaurant reviews and yielded an accuracy of 78.04%.

Braud & Denis presented a framework for identifying implicit discourse relations among words [4]. The authors also illustrated a technique to derive discourse segment representation from vectors of words.

Hajar and Mohammed present a hybrid approach that integrates a corpus-based method with a dictionary-based model for extracting *implicit aspect terms* (IATs) [5]. The hybrid model is then used for supporting naïve Bayes training for the identification of implicit aspects. The naïve classifier is used for evaluating reviews of electronic products and restaurants. The NB model is found to outperform the corpus-based model when used in combination with the proposed hybrid model by the authors.

Wu et al. presented a joint detection model for extracting the triples (target, aspect, sentiment) from a sentence [6]. The proposed model by the authors uses BERT to obtain the initial embedding vector from the aspect–sentence and then applies a bidirectional long short-term memory model for combining the aspect and sentence representations. The dependency relationships between the aspects and the sentences are captured using a graph convolutional network. The performance of the proposed framework is evaluated on two restaurant datasets of SemEval 2015 Task 12 and SemEval 2016 Task 5. The results showed that the model is highly accurate in extracting the sentiments from the targets.

Tang et al. present a mechanism using a dependency graph on a dual-transformer network (named DGEDT) [7]. The mechanism iteratively combines the flat representations discovered from Transformer and graph-based representations explored from the corresponding dependency graph. Specifically, a dual-transformer structure is utilized so that the DGEDT is capable of reinforcement learning from the flat and graph-based representation. Experiments were conducted on five datasets, and the results demonstrate that the proposed DGEDT outperforms all state-of-the-art models.

Hou et al. proposed a graph-ensemble method, GraphMerge, for combining multiple dependency trees to carry out aspect-level sentiment analysis [8]. The proposed method has the unique capability of performing union operations on the edges from different parsers without the requirement of inclusion of any additional parameters in the model. This makes the model less prone to parse error while being computationally lightweight.

Shi et al. presented a contrastive learning method for aspect detection [9]. The model includes two attention modules for representing every segment with word embeddings and aspect embeddings and carries out an effective mapping from one set to another. The aspect detection is further improved using a knowledge distillation method. Entropy filters are introduced to increase learning accuracy during the training. The performance results of the model indicate its high level of accuracy and effectiveness in capturing the aspects.

Tian et al proposed a novel approach for aspect-based sentiment analysis that uses a type-aware graph convolutional network [10]. The mechanism creates an input graph based on the dependency parsing of the input tree. The edges of the graph are created considering both dependency relations and types for the input sentence. For each word, attention to the weights of all such type-aware edges associated with the word is computed. The attention layers are ensembled to carry out comprehensive and robust learning of the aspects. The results demonstrated the effectiveness and accuracy of the proposition.

Wang et al. propose a span-based anti-bias aspect representation learning framework to eliminate sentiment bias [11]. The mechanism uses adversarial learning against the prior sentiments of the aspects to eliminate the sentiment bias. Then, Further, the model orients the distilled opinion candidates with the aspect by span-based dependency modeling for extracting the interpretable opinion terms. The experimental results indicate the effectiveness of the model.

Zhao et al. proposed an end-to-end mechanism for extracting pair-wise aspects and opinion terms [12]. The authors focus on the problem of opinion mining from a perspective of joint term and relation extraction rather than just a formulation of sequence tagging. The proposed method involves multi-task learning based on shared spans, in which extraction of terms is done under the supervision of span boundaries. The model is found to outperform most state-of-the-art techniques.

Liu et al. present a comprehensive survey of various methods and techniques existing in the literature [13]. The authors also provide a comparative analysis of machine learning and deep learning methods used in aspect-based opinion mining.

Yu et al argue that most aspect-based models are poor in exploring the syntactic relations between aspect terms and opinion terms [14]. This problem leads to inconsistencies between the model predictions and the syntactic constraints. The authors propose a framework that implicitly captures the relations between the two tasks, and then based on a global inference method, uncovers the relations between syntactic constraints and aspect terms.

Wu et al. proposed a novel mechanism for extracting pair-wise aspect and opinion terms [15]. The proposition included a syntax fusion encoder that incorporated rich syntactic features, including dependency edges and labels along with the part of speech tags. While forming the aspect-opinion term scoring, both aspect scoring and syntactic scoring are considered. The results showed that the approach is highly effective in extracting aspects and opinion terms.

Xing et al. presented an aspect-aware LSTM model for modeling context [16]. The proposed model has the capability to select important information about the target and can eliminate irrelevant information using its control systems. The authors demonstrated that aspect-aware LSTM can generate more effective contextual vectors than classic LSTM models.

Wang et al. proposed a new learning method that can automatically meta-mine knowledge from multiple past domains [17]. Furthermore, a lifelong learning memory network (L2MN) is designed that incorporates the mined knowledge into its learning process. In the learning process, two types of knowledge are involved, (i) aspect-sentiment attention, and (ii) context-sentiment effect. The model is found to be efficient and accurate in sentiment classification.

Ke et al. proposed a new model called CLASSIC that is based on a contrastive continual learning method [18]. The model can execute both knowledge transfer across tasks as well as knowledge distillation from old tasks to the new task, thereby eliminating the requirement of assignment of task ids during the testing phase. The model is found to be highly efficient as per the results reported by the authors.

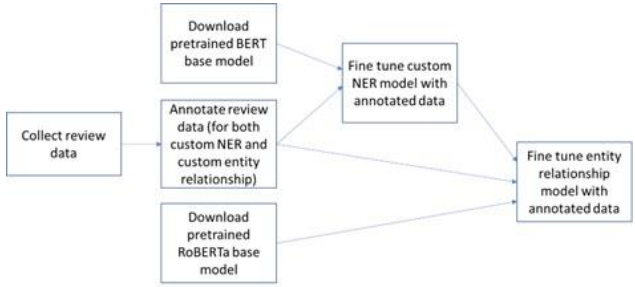

Figure 1. The workflow of the methodology for building a custom relation extraction model.

III. DATA AND METHODOLOGY

For the study at hand, two things were important, first, collecting data and, second, training a model to extract entity relationships. The first task was relatively easy as lots of reviews were available on mobile phones on different websites and using web crawling and web scraping techniques reviews could be collected without much difficulty. The second task was rather somewhat complicated as entity relationship in texts requires lots of manual annotations of words. These annotations involve tagging of words which falls under the task of named entity recognition (NER). Apart from this task, the relationships between entities are also required to be specified. From the context of NLP, for proper model building, lots of annotations are required which itself is a mammoth task. However, in the recent past, a significant amount of work has been done to identify common entities like the person, date, organization, etc. based on annotated texts using artificial neural networks, and the trained network architectures are stored for future works. For the second task, one such model (RoBERTa) was considered as the base model, using which, the custom model was built. For building a custom model, manual annotation of texts was required but the number of such annotations was way less than the number of annotations required had the model been built from the scratch. The custom models were to be built using the concept of transfer learning. The methodology followed in this work is schematically depicted in Figure 1.

A. Data Collection

Extracting aspect-specific opinions from the text is far from a trivial task when the text contains improper English with varying levels of grammatical mistakes. Consumer reviews, in most cases, contain improper grammar and improper usage of punctuation. For this study, the reviews taken under consideration were about mobile phones. The reviews were collected from the Amazon website [19]. A sample review is shown in Figure 2. The review is easily understandable by humans as we can easily neglect grammatical errors. But, when the review is required to be analyzed by a machine to extract aspects and associated opinions, grammar would play an important role. This is because correct *part of speech* (POS) tagging would be important for these extractions but grammatical errors would induce incorrect POS tagging. Training a model from scratch is a very difficult task even with corpora with proper English. Luckily, there are pre-trained models available that are trained with proper English texts having almost no grammatical errors. These models can be fine-tuned as per our requirements and with a limited number of texts, these models can be made adaptable to the current task at hand. This is typically the concept of transfer learning. To achieve this objective, around 200 reviews from the Amazon website were collected. The reviews were collected randomly. The reviews were stored as individual text files after capturing other metadata such as the date of review, review ratings, etc. These text files were required to be annotated manually for fine-tuning the model.

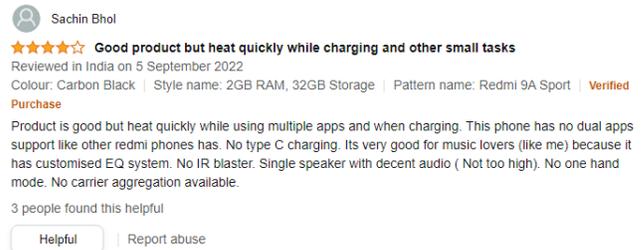

Figure 2. A typical user review of a mobile phone at the Amazon website.

B. Data Annotation

The next task was to open each file individually and annotate words (and phrases) as either an aspect (denoted by ASP tag) or an opinion (denoted by OPI tag). Not every word was required to be annotated. For this purpose, an online tool was used called UBIAI™. This tool gives an easy interface to tag words and associate words with each other representing a relationship between words (or chunks of words). For this task, targeted words were tagged as either ASP or OPI and their relationship was given another tag ASP-OPI. An example of tagging is shown in Figure 3. The tagged outputs were stored as JSON files depicting the words with associated tags. A total of 150 such reviews were tagged manually to fine-tune the pre-trained model.

C. Training the Custom Model

To train the custom model, a pre-trained model was downloaded from the Huggingface website [20]. This model is based on the concept of *transformers*. The downloaded models were 'RoBERTa-base' and a 'BERT-base'. RoBERTa is essentially a BERT model but it is trained with a larger size of the corpus. It does not try to predict the next sentence which is an integral part of the BERT model. Also, it adopts a dynamic masking pattern on the training data resulting in an improved version of basic BERT. It is the base portion of the model that can be used for *transfer learning*, and in this study, the same method is utilized as only a limited number of training data points were available. The training process involved the training of two separate models. The first model was responsible to identify the word (or word chunks) as either aspects or opinions. For example, in Figure 3, the word chunk "great value for money" was required to be predicted as "OPI" whereas "camera" needed to be tagged as "ASP". In that sense, the first model was, basically, a Named Entity Recognition (NER) model. These aspects (ASP) and opinions (OPI) tags were custom tags and hence, pre-trained models were unable to identify them. This is where the annotated texts were supplied as the inputs. However, it was kept in mind that the relationship tags were not supplied to this model. To train the model, the *Spacy* package was utilized and the model was trained using the scripts provided under the 'spacy project'. UBIAI gives the annotation outcome in a format that spacy can handle to create its own binary file that goes into the training process as the required input. Another model using RoBERT-base is also trained to find the relationship between the custom tags.

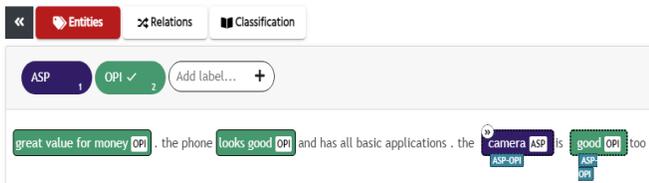

Figure 3. A sample scenario of test annotation using the UBIAI tool

## IV. PERFORMANCE RESULTS

In this section, the performance results of the proposed NER model are presented. The models are implemented, trained, and tested in GPU (Nvidia GEFORCE GTX 1060, 6GB memory). Pytorch version 1.12.2, Spacy natural language processing library is used for designing and training the models.

Figure 4 depicts the training progress with respect to the number of epochs. Three metrics are shown in the figure. The three metrics are *precision* (denoted by *P*), *recall* denoted by *R*, and *F1 score* (denoted by *F*). The performance scores were calculated on the validation data. As depicted in Figure 4, the precision of the model stays near 70% whereas the recall scores were quite less, leading to increased false negative cases. This was expected as the model was fine-tuned with fewer data and the data itself contained improper grammar. Hence, the model failed to detect the correct tags on the validation data for several instances. In terms of the F1 score, the best model yielded an F1 score close to 60% which is at the threshold of acceptance level for industry applications.

Figure 5 shows the performance of the relation extraction model which is the second model trained in this study. The relation extraction model takes input from the NER model as well as the training data to do a classification task in a sequential manner. For the current study, the number of classes was chosen to be one, i.e., only the "ASP-OPI" class was considered. Once the model was trained, raw texts could be sent through the two models to extract the aspects and associated opinions from the text. As can be seen from Figure 4, the performance of the best model was above 60% in terms of its F1 score (at 400 epochs). A sample outcome of these models is presented in Table 1.

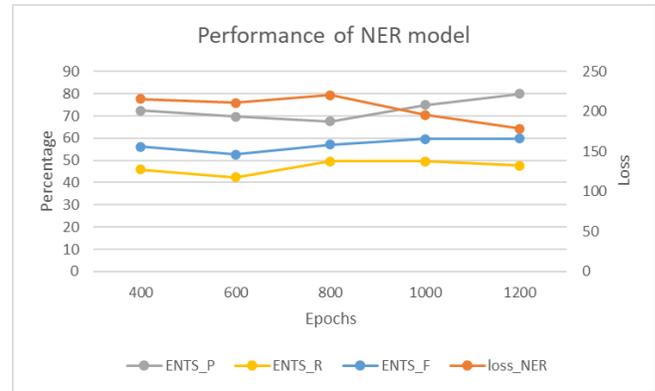

Figure 4. The performance of the NER model in terms of loss, precision, recall, and F1-score

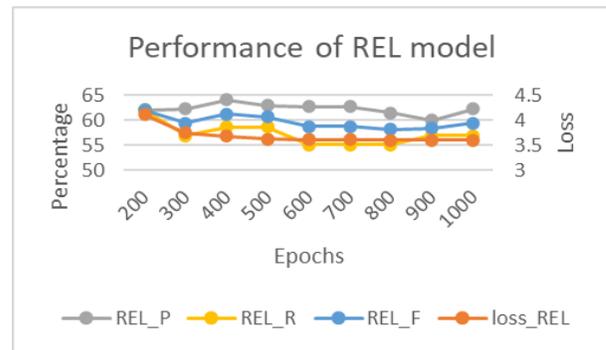

Figure 5. The performance of the REL model in terms of loss, precision, recall, and F1-score

### A. Discussion of the Results:

It can be seen from the above analysis that the relationships could be extracted by the model even though the probability of detection is not very high. Moreover, some of the relationships have appeared between opinions only which is not desirable. However, if the reviews are read carefully, it is apparent that the reviews are containing lots of grammatical errors. With such errors and with limited data points also the model could identify the relationships with decent performance. The model can be used on a larger volume of such texts to create more training data points based on correct predictions and can be retrained based on the new training set. This way, in a semi-supervised manner, the model's performance can be improved to a much better level. This is the next scope of the study and once such a model is created with significantly better performance, analysis of consumer reviews could be done using NLP techniques to gather actionable insights.

TABLE I  SAMPLE OUTPUTS OF THE MODEL- ASPECTS AND OPINIONS EXTRACTED FROM THE REVIEWS

| Sl No | Text | ASP | OPI | Probability (%) |
|---|---|---|---|---|
| 1 | As in this budget we are getting 4 GB RAM and 128 GB internal memory which is quite good combination where other phones are giving 64 GB as their internal storage. | RAM | quite good | 47.9 |
| | | Internal memory | quite good | 63.78 |
| | | Internal storage | quite good | 82.79 |
| 2 | Good features and camera quality is good and powerful battery and nice and auto call recorder option is also available if don't want auto call recorder option manual call recorder option is also available but the problem is we should have click tow time to open any thing other good phone value of money and most recommended | battery | good | 38.72 |
| | | Manual call | nice | 36.81 |
| | | good | nice | 76.90 |
| 3 | Poor screen color, poor camera, wifi also only 2 | Screen color | poor | 74.96 |
| | | camera | poor | 77.20 |
| | | wifi | poor | 51.47 |

## V. CONCLUSION

Consumer reviews hold important information about the product features and their level of acceptance by the end users. However, they fall under unstructured data, and capturing information from unstructured data is way a complex task. With respect to customer reviews, it is critical to understand the aspects on which the review is focused. The associated opinions are also important for understanding the degree of acceptance (or rejection) of the aspects. This paper has presented a framework of aspect-based opinion mining based on the concept of transfer learning. The model has yielded quite satisfactory results in its task of aspect-based opinion mining. Future research includes sentiment analysis of the extracted opinions and improving the performance of the opinion mining framework using advanced approaches like graph convolutional networks and attention-based LSTM networks.